\documentclass[final,1p,times,twocolumn]{elsarticle}

\usepackage{amssymb}
\usepackage{booktabs}
\usepackage{amsmath}
\usepackage{multirow}

\journal{Computers in Biology and Medicine}

\begin{document}

\begin{frontmatter}



\title{echoGAN: Extending the Field of View in Transthoracic Echocardiography Through Conditional GAN-Based Outpainting}

\author[CORTEXVISION]{Matej Gazda}
\author[UNLP,CORTEXVISION]{Jakub Gazda}
\author[MONTREAL,POLYMTL]{Samuel Kadoury}
\author[CORTEXVISION]{Robert Kanasz}
\author[CORTEXVISION]{Peter Drotar}
\affiliation[TUKE]{organization={Technical University of Kosice, Slovakia},
            addressline={Letna 9},
            city={Kosice},
            postcode={040 01},
            country={Slovakia}}

\affiliation[UNLP]{organization={Pavol Jozef Safarik University and Louis Pasteur University Hospital},
            addressline={Srobarova 2},
            city={Kosice},
            postcode={041 90},
            country={Slovakia}}
\affiliation[MONTREAL]{organization={Universite de Montreal},
            addressline={boulevard Édouard-Montpetit},
            city={Montreal},
            postcode={H3T 1J4},
            country={Canada}}

\affiliation[POLYMTL]{organization={Polytechnique Montréal},
            addressline={2500, chemin de Polytechnique},
            city={Montreal},
            postcode={H3T 1J4},
            country={Canada}}

\affiliation[CORTEXVISION]{organization={CortexVision}, 
    ddressline={Turgenevova 33 },
    city={Košice},
    postcode={040 01},
    country={Slovakia}
}

\begin{abstract}
Transthoracic Echocardiography (TTE) is a fundamental, non-invasive diagnostic tool in cardiovascular medicine, enabling detailed visualization of cardiac structures crucial for diagnosing various heart conditions. Despite its widespread use, TTE ultrasound imaging faces inherent limitations, notably the trade-off between field of view (FoV) and resolution. This paper introduces a novel application of conditional Generative Adversarial Networks (cGANs), specifically designed to extend the FoV in TTE ultrasound imaging while maintaining high resolution. Our proposed cGAN architecture, termed echoGAN, demonstrates the capability to generate realistic anatomical structures through outpainting, effectively broadening the viewable area in medical imaging. This advancement has the potential to enhance both automatic and manual ultrasound navigation, offering a more comprehensive view that could significantly reduce the learning curve associated with ultrasound imaging and aid in more accurate diagnoses.  The results confirm that echoGAN reliably reproduce detailed cardiac features, thereby promising a significant step forward in the field of non-invasive cardiac naviagation.
\end{abstract}



\begin{keyword}


medical imaging \sep ultrasound
\end{keyword}

\end{frontmatter}







\section{Introduction}\label{sec1}

Transthoracic Echocardiography (TTE) Ultrasound (USG) is a cornerstone diagnostic tool in cardiovascular medicine, valued for its non-invasive approach and ability to comprehensively assess cardiac anatomy and function \cite{TTE1, TTE3}. It enables detailed visualization of cardiac structures, such as chambers, valves, and myocardium, and is essential in diagnosing conditions like heart failure, valvular heart disease, and cardiomyopathies. One of TTE USG’s key advantages is its capacity for real-time dynamic imaging, allowing clinicians to observe cardiac motion and provide insights beyond what static imaging can offer.

Despite its wide applications, TTE USG has inherent limitations, particularly in the trade-off between field of view (FoV) and resolution. Achieving high resolution in a focused area typically results in a narrower FoV, providing detailed imagery at the cost of a broader view. Conversely, a wider FoV can reduce the resolution, compromising detail for a more comprehensive perspective. This balance between FoV and resolution is a crucial consideration in clinical ultrasound imaging.

Different ultrasound probes are tailored for specific medical purposes. Linear array probes are effective for shallow structures, while curvilinear array probes are suited for deeper tissue imaging. Endocavitary probes excel in specialized fields such as urology and gynecology, offering high-resolution images of internal structures. Sector array probes, often used in cardiac imaging, balance FoV with depth. However, these probes still confront the challenge of balancing FoV, resolution, and penetration depth.

In this context, our research has explored the application of a  conditional Generative Adversarial Networks (cGANs) to extend the FoV in TTE USG imaging while maintaining high resolution. The proposed cGAN architecture, denoted as echoGAN, is capable of generating realistic anatomical structures through outpainting, thereby broadening the viewable area in medical imaging. This approach aims to address the limitations in FoV and resolution trade-offs, potentially enhancing both automatic and manual ultrasound navigation. For automatic navigation \cite{ai_nav}, an expanded FoV can provide a more comprehensive view, potentially improving the efficiency and accuracy of imaging processes, especially in complex diagnostic scenarios. Similar challenges in expanding the FoV have been addressed in related fields, using advanced lens models and flexible array technology to enhance the visual scope and resolution \cite{beam, arr}.

\begin{figure*}
\centering
    \includegraphics[width=0.75    \textwidth]{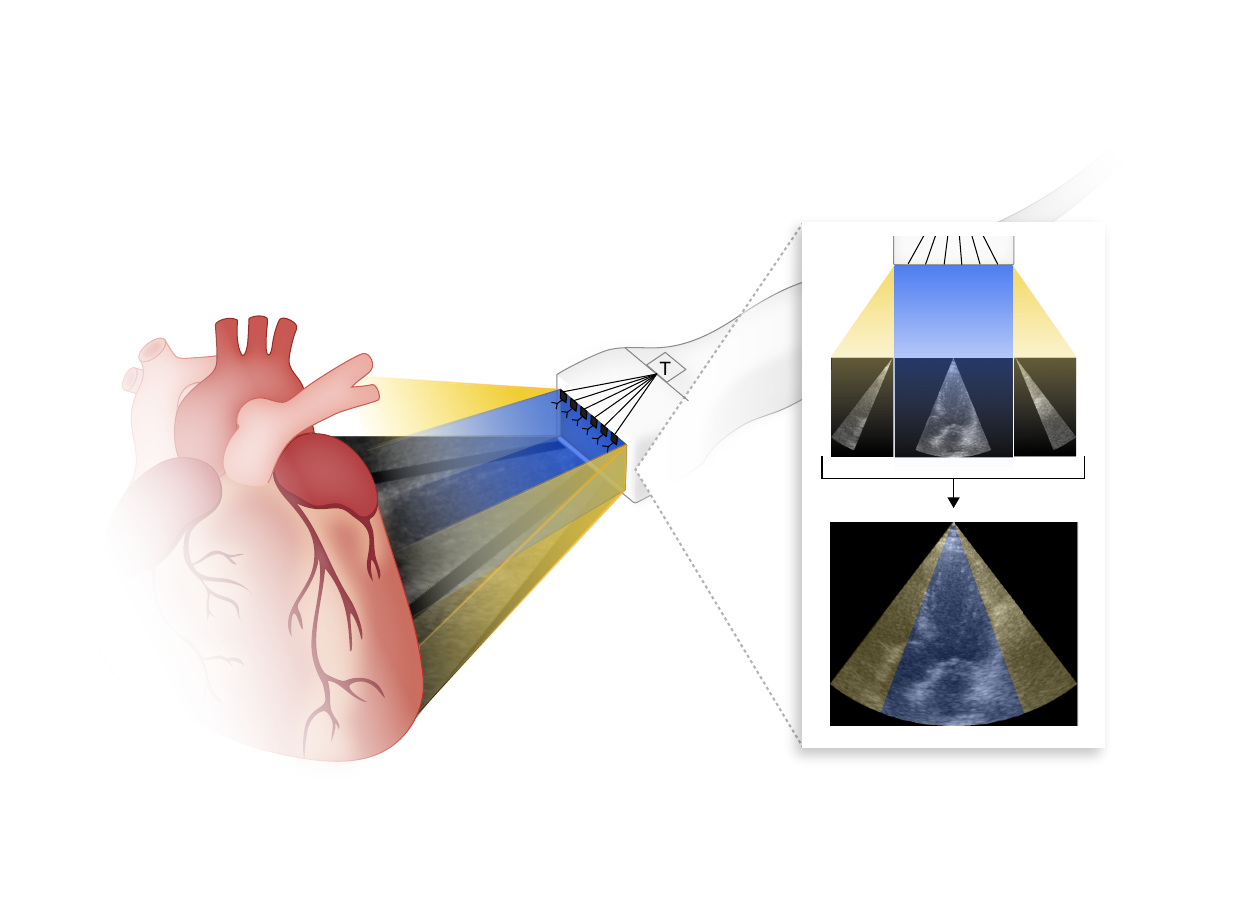}
    \caption{Extending field of view (yellow) vs basic (teal)}
    \label{fig_echogan}
\end{figure*}

Furthermore, this extended FoV could be particularly beneficial for less experienced physicians in manual navigation.The study by Chisholm et al. \cite{chisholm2013focused} and the work of Vinodth et al. \cite{nanjayya2019levels} focus on cardiac ultrasound training, specifically evaluating the duration of training necessary to achieve proficiency in TTE. These studies demonstrate that mastering TTE requires a considerable amount of time. By offering a wider yet detailed view, our cGAN-based approach could reduce the learning curve associated with ultrasound imaging. This process is often not straightforward and demands a high level of dexterity and spatial awareness. Given these challenges, there is a continuous search for technological advancements that can aid in simplifying the navigation process, making echocardiography more accessible and reliable, especially for less experienced practitioners. Moreover, the utilization of semi-synthetic data to enhance training models for medical device segmentation \cite{synt} underscores the potential for augmented datasets to refine AI-driven tools.

Overall, proposed advancement in TTE USG technology through the integration of cGANs significantly contribute to the field of non-invasive cardiac diagnostics. By addressing some of the fundamental limitations of current ultrasound technology, this approach holds the promise of improving navigation processes and in turn patient care in cardiology. The showcase of the novel approach proposed in this paper is depicted in Fig.  \ref{fig_echogan}.

In ultrasound imaging, clinicians have the capability to manipulate the field of view to enhance image resolution. This is achieved through the focusing of the ultrasound beam, which can be narrowed to increase the lateral resolution at specific depths, known as the focal zone. By adjusting the focus and consequently decreasing the FoV, the ultrasound system can provide a more detailed view of the area of interest. This technique is particularly useful in cardiac imaging, where high resolution is critical for accurate diagnosis. The adjustment of the FoV and focus is an integral part of ultrasound knobology, allowing for optimization of image quality based on the clinical requirements of the examination \cite{ng2011resolution}.

In the domain of medical imaging, particularly ultrasound imaging, the task of outpainting introduces a set of unique challenges that are markedly distinct from those encountered in the outpainting of general or natural images \cite{inout,Li_2022_CVPR,rego}. The paramount requirement for anatomical accuracy in medical images necessitates that outpainting algorithms not only generate visually plausible extensions but also ensure these extensions accurately represent the expected anatomy. This is crucial, as any deviation could compromise the clinical utility of the image. Moreover, ultrasound images are characterized by specific textures and patterns that denote various tissues, fluids, and pathological conditions.

The paper is organized as follows. In the subsequent section we  provide description of the proposed approach, detailing echoGan framework which leverages  domain-aware augmentation and conditional GAN architecture. The third section describes the datasets utilized to validate the proposed approach. The fourth section discusses the results obtained from our experiments. Following discussion, the paper concludes with final remarks on the study's findings and implications.


\section{Field of View Outpainting}\label{sec2}

The present study proposed GAN for extensting the FoV in echocardiography to enhance both manual and automatic navigation. From computer vision perspective is FoV extension formulated as outpainting problem. Outpainting seeks for a semantically consistent extension of the input image beyond its available content. The outpainting process involves the GAN generating new pixels and features in a way that seamlessly integrates with the existing content, maintaining the anatomical and contextual coherence of the cardiac structures. 

This approach not only promises to refine navigation accuracy but also to facilitate a more comprehensive understanding of the heart's structure and function. Enhancing the FoV through outpainting reduces navigation difficulty and provides a more comprehensive and detailed view.  The proposed approach aims to address  longstanding challenge of incomplete cardiac visualization during echocardiography.

\subsection{Problem formulation}

Given an input image \( x \in \mathbb{R}^{H \times W} \), where \( H \) and \( W \) represent the height and width of the original image, and a binary mask \( m \in \{0, 1\}^{H \times W} \) that defines the region to be outpainted (with \( 1 \) indicating pixels to be retained and \( 0 \) indicating the masked region), the task is to generate an extended output image \( \hat{y} \in \mathbb{R}^{H \times (W + \Delta W)} \) that fills in the missing content in the masked area while preserving visual and anatomical coherence.

The problem can be expressed as learning a mapping function \( G: (\mathbb{R}^{H \times W} \times \{0, 1\}^{H \times W}) \rightarrow \mathbb{R}^{H \times (W + \Delta W)} \) such that:

\begin{equation}
\hat{y} = G(x \odot (1 - m), m)
\end{equation}

where \( \odot \) denotes the element-wise multiplication operator and \( x \odot (1 - m) \) is the input image with the masked region removed, leaving only the known parts.

The objective is to train \( G \) such that the generated image \( \hat{y} \) is indistinguishable from the true extended image \( y \) under a discriminator \( D \).

\subsection{echoGAN}\label{sec3}

\subsubsection{Domain-aware augmentation}

We propose a novel domain-aware augmentation technique specifically designed for ultrasound imaging. This augmentation simulates the cone-shaped FOV typical in ultrasound by masking the side regions of the cone. The augmented mask enforces the generator to reconstruct the occluded areas.

For ultrasound image $x$, the cone-like FOV is defined by a binary mask $ m_{\text{cone}} \in \{0, 1\}^{H \times W} $, where
\begin{equation}
m_{\text{cone}}(i, j) = 
\begin{cases} 
1, & \text{if } \tan^{-1} \left(\frac{|j - c_x|}{H - i}\right) \leq \frac{\theta}{2}, \\
0, & \text{otherwise}.
\end{cases}
\end{equation}
Here, $ m_{\text{cone}}(i, j) = 1 $ for pixels inside the cone-like region and $ m_{\text{cone}}(i, j) = 0 $ for those outside the cone (cut-out regions). The cone is parameterized by the center of the cone $(c_x, c_y)$ and the angular spread  $\theta$.

The augmentation simulates missing information by masking the side regions of the cone, requiring the generator to infer and reconstruct the missing areas by utilising augmented mask $m_{\text{augment}} \in \{0, 1\}^{H \times W}$ defined as 

\begin{equation}
m_{\text{augment}}(i, j) = 
\begin{cases} 
1, & \text{if } \tan^{-1} \left(\frac{|j - c_x|}{H - i}\right) \leq \frac{\theta_{\text{shrink}}}{2}, \\
0, & \text{otherwise}.
\end{cases}
\end{equation}
where \( \theta_{\text{shrink}} < \theta \) controls the extent of the occlusion on the sides.

Then, augmented input image is defined as
\begin{equation}
z_{\text{aug}} = \{x \odot m_{\text{augment}}, m_{\text{augment}}\},
\end{equation}
where $x \odot m_{\text{augment}}$ represents ultrasound masked image retaining only the visible region.

Finally, the generator \( G \) takes the augmented input \( z_{\text{aug}} \) and produces the extended image \( \hat{y} \):
\begin{equation}
\hat{y} = G(z_{\text{aug}}).
\end{equation}

\subsubsection{Network architecture and training}

The proposed GAN architecture  consists of two neural networks : generator and discriminator.  As a generator we employed
    U-Net architecture, as introduced by Ronneberger et al. \cite{ronneberger2015u}. It features two input channels: one for the masked image and the other for the binary mask, which indicates the regions for outpainting. It is configured with four downsampling layers, each characterized by specific kernel sizes [7, 5, 5, 5] and strides [1, 2, 2, 2], optimizing the network's capacity for feature extraction and image reconstruction within the designated areas. 

The Discriminator employs a conventional Convolutional Neural Network (CNN) design, structured with a sequence of channels [1, 32, 64, 128, 128] and uniform strides [1, 2, 2, 2] across layers, with a kernel size consistently set to 3. This configuration enables effective differentiation between generated and authentic images, ensuring the generative model's outputs closely mimic the target distribution.

\begin{figure}[h!]
    \centering
    \includegraphics[trim={0 5cm 0 2cm},clip,scale=0.7]{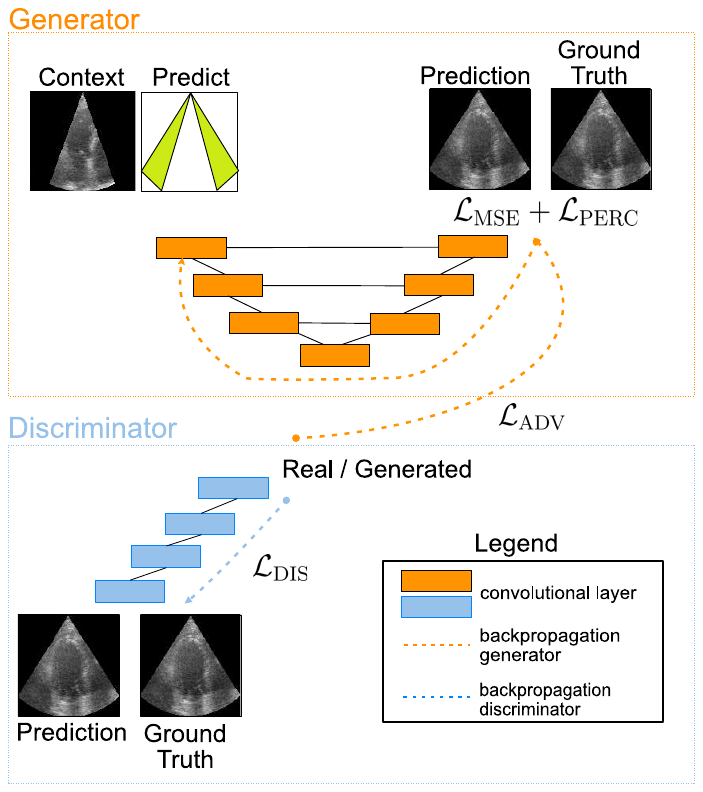}
    \caption{Proposed echoGAN Architecture}
    \label{proposed_nn}
\end{figure}

This network, depicted on Fig. \ref{proposed_nn}, is trained by combination of two different loss functions: adversarial loss function and  learned perceptual image patch similarity. The adversarial loss consists of two components: one for the discriminator ($\mathcal{L}_D$) and one for the generator ($\mathcal{L}_G$). The discriminator loss $\mathcal{L}_D$ is designed to correctly distinguish between real and fake images, while the generator loss $\mathcal{L}_G$ aims to fool the discriminator into classifying fake images as real. The adversarial loss for the discriminator can be formulated as:

\begin{equation}
\mathcal{L}_D = -\mathbb{E}_{x \sim p_{\text{data}}(x)}[\log D(x)] - \mathbb{E}_{x \sim p_{\text{data}}(x), m}[\log (1 - D(G(x \odot (1 - m), m)))]
\end{equation}

where \(x\) represents the real images sampled from the data distribution, \(m\) is the mask, and \(x \odot (1 - m)\) denotes the image with the masked area removed. Generator loss is then denoted as
\begin{equation}
\mathcal{L}_G = -\mathbb{E}_{x \sim p_{\text{data}}(x), m}[\log D(G(x \odot (1 - m), m))].
\end{equation}

The Learned Perceptual Image Patch Similarity (LPIPS) metric evaluates the perceptual similarity between two images in a way that is more aligned with human visual perception than traditional metrics like MSE. The general form of the LPIPS equation involves computing differences in deep feature representations extracted from the two images being compared, then weighting and summing these differences to produce a final similarity score. LPIPS is defined as 
\begin{equation}
\mathcal{L}_{\text{LPIPS}} = \sum_{l} w_l \cdot \frac{1}{H_l W_l C_l} \sum_{h,w,c} \| \phi_l(I^1)_{h,w,c} - \phi_l(I^2)_{h,w,c} \|_2^2
\end{equation}

where $l$ is the layer of the deep feature extractor, $H$ accounts for height, $W$ for width, $C$ for number of channels, $\phi$ is the featured extractor (pretrained VGG16).

Final loss for generation is then a summation of both losses
\begin{equation}
\mathcal{L} = \mathcal{L}_{\mathrm{G}} + \mathcal{L}_{\mathrm{LPIPS}}.
\end{equation}

The objective is to find the optimal generator \( G^* \) that minimizes the combined loss \( \mathcal{L} \) while simultaneously training the discriminator \( D \) to maximize \( \mathcal{L}_D \):

\begin{equation}
G^* = \arg \min_G \max_D \left[ \mathcal{L}_D + \mathcal{L} \right].
\end{equation}

This formulation ensures that the generator produces realistic outpainted images that closely resemble true TTE ultrasound images.

\section{Data}\label{sec4}
We utilised two datasets in this paper: CAMUS dataset \cite{camus} collected under the CAMUS project \cite{camus} at the University Hospital of St. Étienne and EchoNet Dynamic echocardiography dataset \cite{ouyang2019echonet}.

CAMUS  dataset was designed to facilitate measurements of Left Ventricular Ejection Fraction (LVEF). For each patient, 2D apical four-chamber and two-chamber view sequences were recorded. These sequences were extracted from video recordings, yielding an average of 20 images per patient (minimum: 10, maximum: 42). In total, the dataset comprises 9,964 images, of which 7,917 images from 400 patients were allocated for training a generative adversarial network. To prevent data leakage, all data from a given patient were confined to either the training set or the testing set, ensuring no overlap.

To further validate the echoGAN, we incorporated  also the EchoNet Dynamic, consisting of 10,030 apical four-chamber echocardiography videos obtained from routine clinical care at Stanford University Hospital between 2016 and 2018. Each video provides a visualization of the heart from various angles and positions using different image acquisition techniques. To standardize the dataset, all videos were cropped and masked to exclude textual information and irrelevant content outside the scanning sector. The resulting frames were subsequently downsampled using cubic interpolation to a uniform resolution of 112 × 112 pixels.

Due to the absence of US data in raw format within the datasets, our analysis is confined to the final, scan-converted images. A similar approach might be implemented in raw data as well. We segmented the cone from the image data through the application of conventional image processing techniques. Initially, we located the first non-zero pixel from the top as a means to identify the tip of the probe. Subsequently, we delineated the furthest left and right boundaries emanating from the probe to the edge of the image. By determining the angular relationship between these boundaries, we systematically removed segments measuring 15, 23, 30, and 40 degrees from each side in a symmetrical fashion. The resulting masks, corresponding to the excised segments, were preserved for further analysis.

As for the preprocessing during the training, we just scaled the intensity into [0, 1] interval. 

\section{Results}\label{sec5}

To quantitatively assess the quality of generated images Fréchet Inception Distance  \cite{dowson1982frechet} was employed. FID aims to capture the perceptual similarity between real and generated images by considering their feature representations extracted from a pre-trained Inception CNN. FID calculates the Fréchet distance between multivariate Gaussian distributions fitted to the feature representations of real and generated images. Formally is FID defined as 
\begin{equation}
FID = ||\mu_{\mathrm{real}} - \mu_{\mathrm{gen}}||^2 + Tr(\sum{\mathrm{real}} + \sum{\mathrm{gen}} - 2(\sum{\mathrm{real}} \sum{\mathrm{gen}})^{1/2}),
\end{equation}

where $\mu$ is mean and $\sum$ is covariance of of the feature vectors for the real and generated images respectively \cite{dowson1982frechet}. $Tr$ is the trace. A lower FID indicates better similarity between the distributions, suggesting higher image quality and fidelity.

\begin{figure}
\centering
    \includegraphics[width=0.35\textwidth]{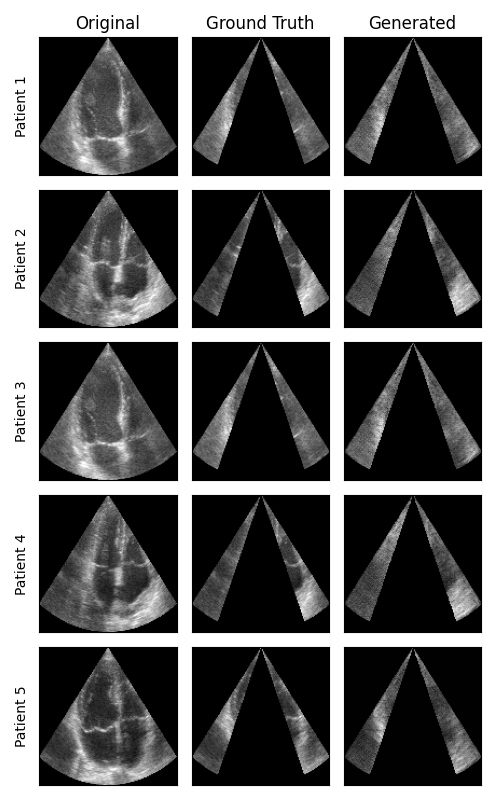}
    \caption{Several examples from different patients on ground truth image and outpainted image.}
    \label{fig:outpaint}
\end{figure}

In Fig. \ref{fig:outpaint}, we provide visual samples of outpainted images generated by the echoGAN. These examples showcase the network's ability to extrapolate relevant cardiac features beyond the visible boundaries, demonstrating the ability of the proposed GAN architecture to enhance the overall imaging capabilities in echocardiography. The visual representations serve to illustrate the effectiveness of the echoGAN in creating realistic and contextually coherent extensions of the original ultrasound images, thereby contributing to the advancement of FoV extension techniques in cardiac imaging applications.  

To quantitatively measure the quality of outpainted images we utilised Fréchet Inception Distance (FID). The average FID for outpainted images, when using 30 degree wide cut for inference was equal to 122.08. 

\subsection{Different size of extended view}

\begin{figure*}
\centering
    \includegraphics[trim={0 0cm 0 3cm},width=1\textwidth]{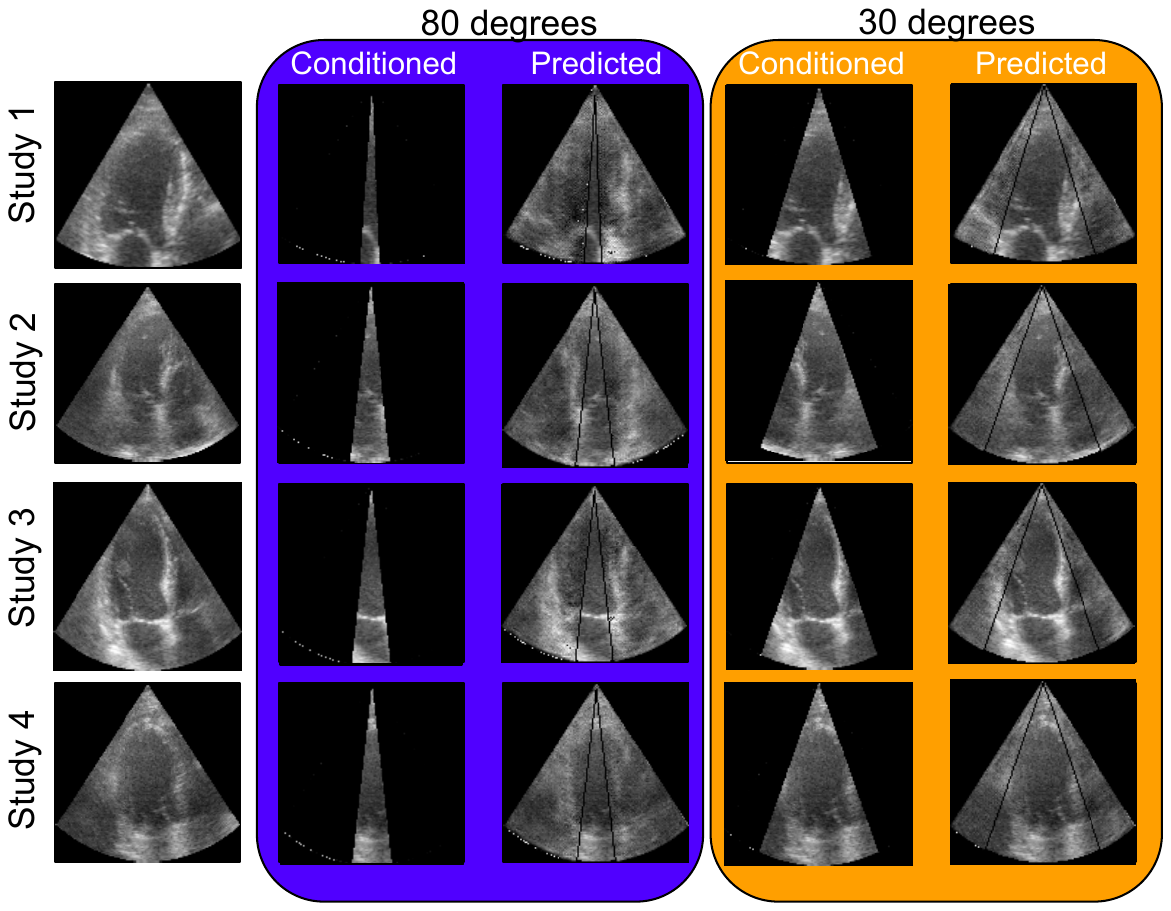}
    \caption{Extending field of view (yellow) vs basic (teal) for 30 and 80 degree outpainting.}
    \label{fig:figure_all_outpaints}
\end{figure*}

We evaluated the influence of the training cone size on the quality of outpainted sections in ultrasound images. Figure \ref{fig:figure_all_outpaints} illustrates two distinct scenarios: the first utilizes a narrower 10-degree cone cut as the inference input, with the echoGAN algorithm tasked with generating outpainted extensions that span 40 degrees on both sides; the second employs a 60-degree cone cut for inference, with outpainting applied to 15-degree segments on each side. Notably, the generated images appear highly realistic, even when echoGAN is interfaced with only a 10-degree cut. The algorithm demonstrates the ability to reproduce cardiac muscle structures with high fidelity. This is particularly evident in the first, second, and fourth rows, where echoGAN successfully outpaints a complete right atrium; that is an encouraging result given that the right atrium was not included in the 10-degree cut, suggesting that echoGAN can reconstruct it entirely. In the case of the left ventricle and left atrium, we observe that echoGAN is capable of outpainting the missing boundary parts for the 30-degree cut and can significantly reconstruct substantial portions of these structures for the 10-degree cut. Again, to quantitatively evaluate the quality of outpainted USG images, we measure FID for different sizes of outpainted regions. The results are provided in Tab. \ref{tab_fid}. The 30-degree cut means that echoGAN generates 15 degrees on both sides. Similarly, in the case of 80 degrees, echoGAN generates 40 degrees on both sides, using only a very narrow 10-degree cone as an input.  


\begin{table*}[h]
\hspace*{-2.5cm}
\centering
\begin{tabular}{c@{}lcccccccc@{}}
\toprule
\textbf{Method} & \textbf{Metric} & \multicolumn{4}{c}{\textbf{CAMUS}} & \multicolumn{4}{c}{\textbf{EchoNet Dynamic}} \\
\cmidrule(lr){3-6} \cmidrule(lr){7-10}
 & & \textbf{Cut 30} & \textbf{Cut 46} & \textbf{Cut 60} & \textbf{Cut 80} & \textbf{Cut 30} & \textbf{Cut 46} & \textbf{Cut 60} & \textbf{Cut 80} \\
\midrule
\multirow{6}{*}{echoGAN} & \textbf{MSE $\downarrow$}  & 0.004893 & 0.008215 & 0.00173 & 0.00124 & 0.01057 & ... & 0.01064 & 0.0175 \\
                         & \textbf{L1 $\downarrow$}  & 0.039 & 0.05033 & 0.063 & 0.06114 & 0.05719 & ... & 0.05653 & 0.08762 \\
                         & \textbf{FID $\downarrow$}  & 122.08 & 143.83 & 145.74 & 149.63 & ... & ... & ... & ... \\
                         & \textbf{LPIPS $\downarrow$} & 0.05304 & 0.09 & 0.126 & 0.1211 & 0.1583 & ... & 0.1567 & 0.2174 \\
                         & \textbf{PSNR $\uparrow$}  & 23.91 & 21.46 & 19.77 & 19.95 & 20.15 & ... & 20.13 & 17.89\\
                         & \textbf{SSIM $\downarrow$}  & 0.2749 & 0.3088 & 0.4707 & 0.4596 & 0.4799 & ... & 0.4815 & 0.5881 \\
\midrule
\multirow{6}{*}{CoMoDoGAN \cite{zhao2021large} } & \textbf{MSE $\downarrow$}  & 0.1376 & 0.16 & 0.178 & 0.1720 & ... & ... & ... & ... \\
                           & \textbf{L1 $\downarrow$}  & 0.32 & 0.368 & 0.4 & 0.3902 & ... & ... & ... & ... \\
                           & \textbf{FID $\downarrow$}  & 109.59 & 158.68 & 208.59 & 194.03 & ... & ... & ... & ... \\
                           & \textbf{LPIPS $\downarrow$} & 0.1642 & 0.2068 & 0.2732 & 0.2584 & ... & ... & ... & ... \\
                           & \textbf{PSNR $\uparrow$}  & 11.589 & 9.3082 & 7.5371 & 7.7864 & ... & ... & ... & ... \\
                           & \textbf{SSIM $\downarrow$}  & 0.5674 & 0.7043 & 0.7973 & 0.7974 & ... & ... & ... & ... \\
\bottomrule
\end{tabular}
\caption{The results for different sizes of outpainted area across CAMUS and EchoNet datasets.}
\label{tab_fid}
\end{table*}

As expected, the smaller FID values indicating better similarity are achieved for the cases when the broader cut is used as an input for network inference.

\subsection{Right ventricle area}

The majority of diagnostic, prognostic, and monitoring efforts in cardiology traditionally emphasize the analysis of the left heart due to its direct role in systemic circulation. However, comprehensive echocardiographic assessment necessitates detailed examination of the entire heart, including the right ventricle (RV), which plays a pivotal role in pulmonary circulation and systemic venous return \cite{echo_rv}. During standard echocardiography, capturing the complete anatomy of the right heart can be challenging. The RV, in particular, may be partially or entirely outside the ultrasound beam's FoV, limiting the accuracy and completeness of the examination. Despite the primary focus on the left heart in clinical practice, quantification of right heart parameters is crucial in various clinical scenarios.

Our experiments are specifically aimed at the quantification of the RV area from both original and outpainted echocardiograms. The rationale for focusing on the RV area is influenced by the established practice in cardiology where, for instance, the measurement of left ventricular ejection fraction incorporates the left ventricular area as a crucial parameter in its evaluation \cite{npj_lvef}. Drawing inspiration from such methodologies, we hypothesize that if the RV areas derived from generated (outpainted) and real echocardiograms are similar, it would substantiate the utility and validity of the generated images for clinical purposes.


Moreover, accurate assessment of RV size and function is paramount in conditions such as Pulmonary Hypertension, various Cardiomyopathies, and Right Ventricular Infarction, where the right heart's performance directly influences patient outcomes\cite{echo_rv}.

\begin{figure}[h!]
    \begin{minipage}[t]{0.45\textwidth}
        \includegraphics[width=\textwidth]{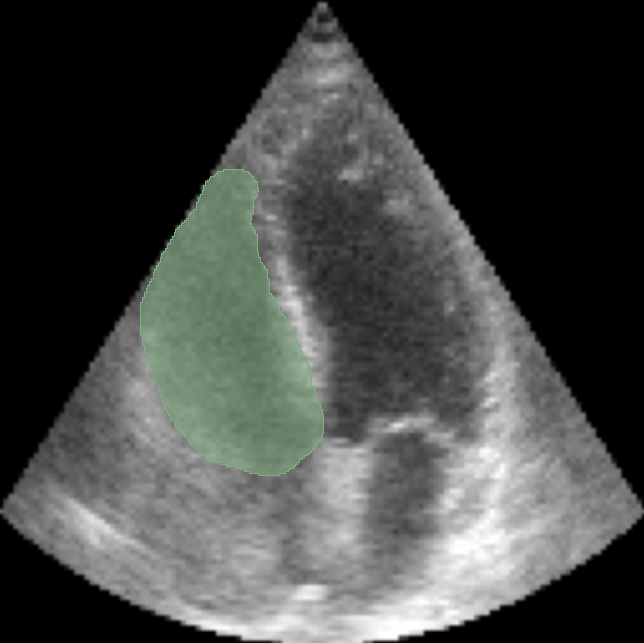}
         \caption{Segmentation over Right Ventricle for GT 4CH view}
        \label{fig:gt_seg}
    \end{minipage}
    \hfill
    \begin{minipage}[t]{0.45\textwidth}
        \includegraphics[width=\textwidth]{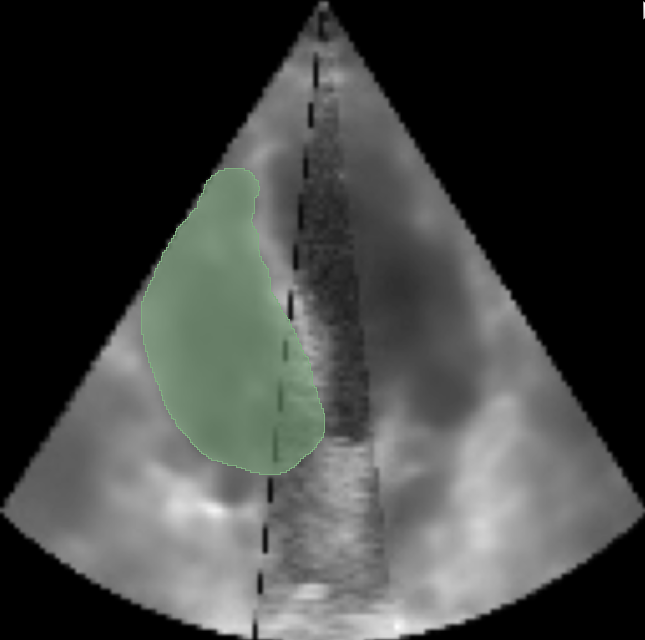}
        \caption{Segmentation of outpainted RV with 80 degrees cut for the same 4CH view}
        \label{fig:predicted_seg}
    \end{minipage}
\end{figure}

In our experimental setup, both the original and the artificially generated echocardiograms were manually annotated to obtain segmentation masks. This process allowed for precise comparison of RV volumes, assessing the efficacy and accuracy of GAN-based outpainting in reproducing anatomically coherent extensions of the cardiac structures. By comparison of the RV volumes derived from the conventional and outpainted echocardiograms, this experiment aims to validate the potential of echoGANs in overcoming the intrinsic limitations of current echocardiography practices. Comparison of segmentation for RV on original and on outpainted image is in Fig. \ref{fig:gt_seg} and Fig. \ref{fig:predicted_seg}.

To statistically evaluate the difference between the real and AI-generated right ventricle area measurements, we first assessed whether each group of measurements conformed to a normal distribution. This preliminary step involved applying the Shapiro-Wilk test \cite{shapiro}, a widely accepted method for testing normality. The results of the Shapiro-Wilk test indicated that the data do not follow a normal distribution, thereby necessitating the selection of a non-parametric statistical test for further analysis. Consequently, we opted for the permutation test, a robust non-parametric approach that does not rely on assumptions of normality or symmetry. This test was employed to examine our hypothesis that there is no significant statistical difference between the areas determined from the AI-generated and real echocardiography data. The choice of the permutation test was appropriate given the non-normal distribution of our data, allowing us to conduct a reliable comparison of the paired differences between the two groups.  The results of this test confirmed the null hypothesis ($T=8.804$, $p=0.762$), indicating that there is no statistically significant difference between the RV areas measured in the original echocardiographic images and those derived from images extended through echoGAN outpainting. This finding validates the accuracy and effectiveness of the GAN-based outpainting technique in echocardiography. The ability of GANs to generate anatomically coherent extensions of the cardiac structures without compromising the integrity of the images suggests that this approach can reliably extend the field of view in echocardiographic navigation. The statistical analysis was conducted using the Python programming language, utilizing the SciPy library.





\section{Discussion}
One of the main limitations of our study is limited data. The dataset used, while extensive, may not fully capture the diversity of cardiac pathologies or patient demographics typically encountered in broader clinical practice. This limitation could affect the generalizability of our findings, as the performance of the echoGAN might vary with different or more heterogeneous datasets. Moreover, while the echoGAN architecture demonstrates the ability to generate anatomically plausible images, there remains a risk of generating artifacts, especially in more complex or ambiguous cardiac regions. These potential inaccuracies necessitate further refinement of the model and algorithms to ensure reliability and clinical utility.

Additionally, a prevalent limitation of publicly available datasets is their focus on specific traditional views like apical two-chamber (A2C) and apical four-chamber (A4C), while neglecting other standard views such as parasternal (parasternal long axis (PLAX), parasternal short axis (PSAX)) and subcostal views (subcostal 4CH and subcostal short axis). Moreover, these datasets completely omit images captured from non-standard views, which are often not even recorded as they are primarily used for navigation purposes.

To address this limitation, future work could incorporate shape priors derived from medical imaging modalities that capture comprehensive chest information, such as CT or MRI. Alternatively, the dataset could be expanded with images from non-standard views generated through real-time ultrasound simulation solutions from CTs or MRIs, as demonstrated in \cite{shams2008real}. In addition to physics-based simulators, unpaired image-to-image translation methods like mUNIT \cite{huang2018multimodal} or CycleGAN \cite{zhu2017unpaired} could be employed. While paired image-to-image translation might yield superior results, the required data is rarely available.

Directly outpainting ultrasound views using scanline data could be a faster and more precise approach, provided that access to the probe hardware and controls were available, which is currently not feasible. However, oversampling the area near the probe and heavily extrapolating at the end of the scanning volume could introduce errors. Should this capability become available, it would offer a promising avenue for further exploration.

\section{Conclusion}\label{sec13}

This paper introduced echoGAN, an innovative approach leveraging conditional Generative Adversarial Networks (cGANs) to extend the field of view (FoV) in Transthoracic Echocardiography (TTE) ultrasound imaging. By addressing the inherent trade-offs between FoV and resolution in standard ultrasound imaging, echoGAN is capable of producing realistic outpainted regions, enhancing the visualization of cardiac structures while maintaining high-resolution detail. This capability can significantly aid both manual and automated ultrasound navigation, reducing the learning curve for less experienced operators and facilitating more accurate diagnoses.

Quantitative analysis demonstrated the robustness of echoGAN, with a Fréchet Inception Distance (FID) of 122.08 for a 30-degree outpainting scenario and slightly higher FID values for larger outpainted regions, such as 143.83 for a 46-degree cut and 149.63 for an 80-degree cut. These results indicate that echoGAN effectively maintains anatomical coherence and quality even as the scope of outpainting increases. Clinical validation through the comparison of right ventricular (RV) area measurements in original and outpainted images further reinforced the reliability of this approach, with no statistically significant differences found, as verified by the Wilcoxon signed-rank test.

While echoGAN shows significant potential, limitations include the need for more diverse datasets and improvements in generating artifact-free extensions, particularly in complex anatomical areas. Future work should explore integrating shape priors from CT or MRI scans and include non-standard cardiac views to enhance anatomical fidelity.

\subsubsection*{Acknowledgment}
Funded by the EU NextGenerationEU through the Recovery and Resilience Plan for Slovakia under the project No. 09I03-03-V04-00394 and project No. 09I03-03-V04-00639.

(Part of the) Research results was obtained using the computational resources procured in the
national project National competence centre for high performance computing (project code:
311070AKF2) funded by European Regional Development Fund, EU Structural Funds Informatization
of society, Operational Program Integrated Infrastructure



\bibliographystyle{iopart-num}
\bibliography{main.bib}

\end{document}